\documentclass[10pt,twocolumn,letterpaper]{article}

\usepackage{cvpr}

\usepackage{epsfig}
\usepackage{amsmath}
\usepackage{amssymb}
\usepackage{subfigure}
\usepackage{amsthm}
\usepackage{amsfonts}
\usepackage{mathrsfs}
\usepackage{booktabs}
\usepackage{multirow, eucal}
\usepackage{helvet}
\usepackage{courier}
\usepackage{bm}
\usepackage{bbm}
\usepackage{graphicx}
\usepackage{color}
\usepackage{url}
\usepackage{booktabs}
\usepackage{soul}
\usepackage[vlined,ruled,linesnumbered]{algorithm2e}
\usepackage{cite}

\usepackage[pagebackref=true,breaklinks=true,letterpaper=true,colorlinks,bookmarks=false]{hyperref}
\usepackage{caption}
\usepackage{tabu}
\usepackage[pagebackref=true,breaklinks=true,letterpaper=true,colorlinks,bookmarks=false]{hyperref}
\usepackage{arydshln}

\def\cL{{\cal L}}

\def\mL{{\mathcal L}}

\def\btheta{\boldsymbol{\theta}}

\def\bx{\bm{x}}
\def\by{\bm{y}}
\def\bQ{\bm{Q}}
\def\bg{\bm{g}}

 \cvprfinalcopy %

\ifcvprfinal\fi
\begin{document}

\title{Training Quantized Neural Networks with 
a
Full-precision Auxiliary Module}
\author{Bohan Zhuang$^1$ ~~ ~ Lingqiao Liu$^1$ ~ ~~ 
Mingkui Tan$^2$  ~~ ~ Chunhua Shen$^1$\thanks{
Corresponding author, e-mail: $\sf chunhua.shen@adelaide.edu.au$} ~~ ~  Ian Reid$^1$
\\
	$^1$The University of Adelaide ~~ ~
	$^2$South China University of Technology
}

\maketitle

\begin{abstract}

In this paper, we seek to tackle a challenge in training low-precision networks: the notorious difficulty in propagating gradient through a low-precision network due to the non-differentiable quantization function. We propose a solution by training the low-precision network with a full-precision auxiliary module.  Specifically, 
during training, we construct a mix-precision network by augmenting the original low-precision network with the full precision auxiliary module. Then the augmented mix-precision network and the low-precision network are jointly optimized.
This strategy creates additional full-precision routes to update the parameters of the low-precision model, thus making the gradient back-propagates more easily.
At the inference time, we discard the auxiliary module without introducing any computational complexity to the low-precision network.
We evaluate the proposed method on image classification and object detection over various quantization approaches and show consistent performance increase. In particular, we achieve near lossless performance to the full-precision model by using a 4-bit detector, which is of great practical value.
\end{abstract}

\section{Introduction}
Deep neural networks (DNNs) have made great strides in many computer vision tasks such as image classification~\cite{krizhevsky2012imagenet, he2016deep}, segmentation~\cite{girshick2014rich, he2017mask} and detection~\cite{ren2015faster, redmon2016you}. Even though deep and/or wide models can achieve promising accuracy, their huge computational complexity makes them incompatible with energy constrained devices which usually have limited memory bandwidth and computational power. This has motivated the community 
to design energy-efficient models, often based on quantized precision, aiming not to sacrifice accuracy relative to the full-precision models. In this paper, we propose to improve the training of the low-precision networks.

The core challenge for quantization is the non-differentiability of the discrete quantizer. As a result, we cannot directly optimize the discretised network with stochastic gradient descent. And the
current solutions can be divided into two categories. The first category is to employ a surrogate of gradient. 
The most commonly used approach is the straight-through estimator (STE)~\cite{bengio2013estimating}. Some recent works have been proposed to relax the discrete quantizer to be continuous for gradient-based optimization~\cite{bai2018proxquant, louizos2018relaxed}. Even though the discontinuity of the discretization operation during training can be partly solved by smoothing it appropriately, some important information may still be missed due to the approximation which can lead to an undesirable drop in accuracy.
The second category is to seeking guidance from a full-precision model for discretised network training. For example, new training strategies such as knowledge distillation~\cite{zhuang2018towards, mishra2018apprentice, zhuang2019effective} have been proposed to learn a low-precision student network by distilling knowledge from a full-precision teacher network. %

\begin{figure*}[h]
  \centering
  \resizebox{0.7580\linewidth}{!}
  {\includegraphics[width=1\textwidth]{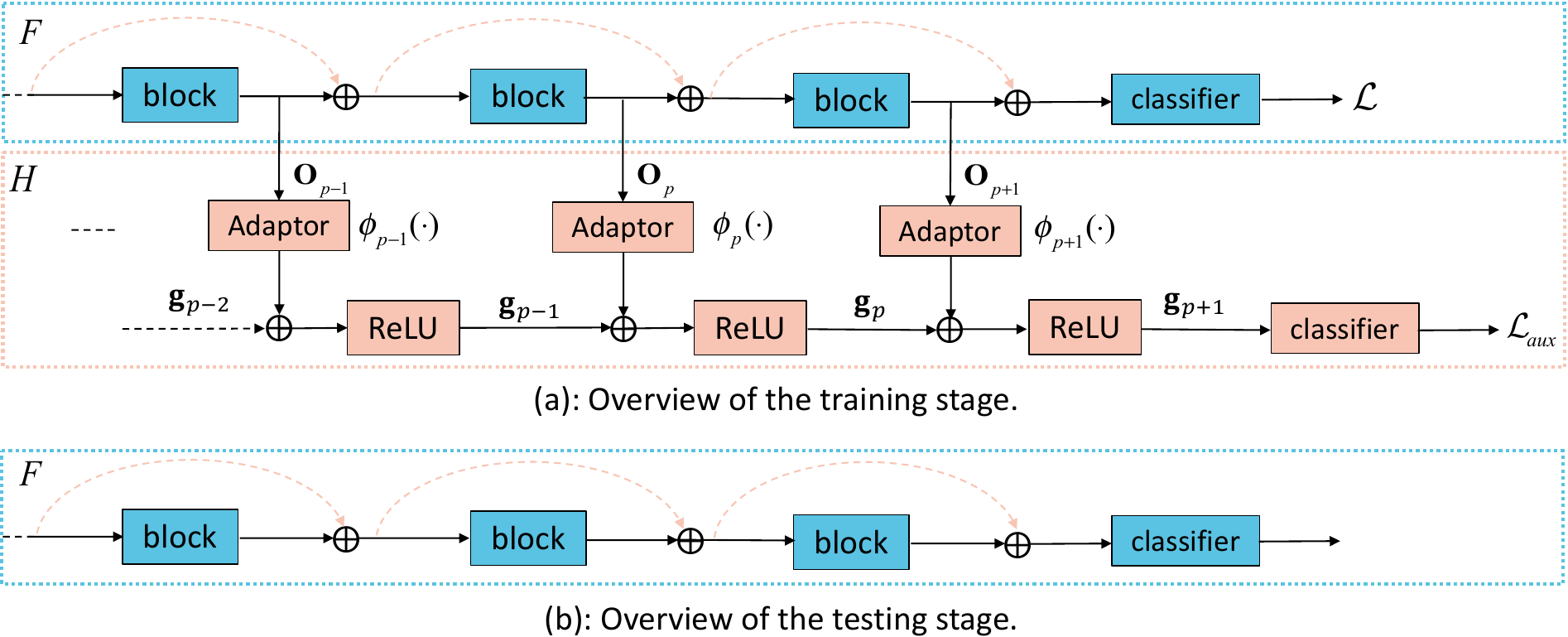}}
  \caption{An overview of the proposed framework. Blue color represents low-precision operations while pink color denotes full-precision operations. 
  During training, the full-precision $H$ is connected to the quantized $F$ to form a mixed-precision network $F\circ H$. Then the mixed-precision network and the low-precision $F$ are jointly optimized with weight sharing.
  It is worth noting that only the learnt quantized network $F$ is used during testing.}
  \label{fig:framework}
\end{figure*}

Our method falls into the second category and our approach is based on the idea of sharing parameters between a mixed-precision (partially fully-precision) model and a low-precision model. 
Specifically, our method constructs a full-precision auxiliary module which connects to multiple layers of a low-precision model (see Fig.~\ref{fig:framework}).
During training, the low-precision network and the full-precision auxiliary module will be combined to form an augmented mixed-precision network. Then, the mixed-precision network and the low-precision model are jointly optimized. Since the parameters from the low-precision model are shared, they can receive gradient from both full-precision connections and low-precision connections. Consequently, the parameters of low-precision model can be updated via two routes, and this can overcome the gradient propagation difficulty due to the discontinuity of the quantizer. Note that only the low-precision network will be utilized for inference and therefore there is no additional complexity introduced at the test stage.

In addition to image classification, we further extend the proposed approach to building quantized networks for object detection. Building low-precision networks for object detection is more challenging since detection needs the network outputs richer information, such as locations of bounding boxes. 
There has been several works in literature to address the quantized object detectors \cite{wei2018quantization, li2019fully, jacob2017quantization}. However, there still exists a significant performance drop of 4-bit or lower-precision quantized detectors comparing to their full-precision counterpart. We apply our techniques to train a 4-bit RetinaNet \cite{lin2017focal} detector and further propose a modification to RetinaNet to better accommodate the quantization design. Through extensive experiments on the COCO benchmark, we show that our 4-bit models can achieve near lossless performance comparing to the full-precision model,  which has a significant value in practice.

Our contributions can be summarized as follows:
\begin{itemize}

\itemsep -0.125cm

     \item{We propose a new training method to account for the non-differentiability of the quantization operator in a low-precision network. Our method can lead to more accurate low-precision model without increasing the model complexity at the testing stage. }

	\item{We apply our learning approach and propose a new design modification to build a 4-bit quantized object detection which achieves comparable performance to its full-precision counterpart.}

\end{itemize}

\subsection{Related work}
	
\noindent\textbf{Network quantization.}  
Quantized network represents the weights and activations with very low precision, thus yielding highly compact DNN models compared to their floating-point counterparts. Moreover, the convolution operations can be efficiently computed via bitwise operations.
Quantization can be categorized into fixed-point quantization and binary neural networks (BNNs), in which fixed-point quantization can also be divided into uniform and non-uniform. Uniform approaches~\cite{zhou2016dorefa, zhuang2018towards, jung2019learning} design quantizers with a constant quantization step. To reduce the quantization error, non-uniform strategies~\cite{Cai_2017_CVPR, zhang2018lq} propose to learn the quantization intervals by jointly optimizing parameters and quantizers. 
A fundamental problem of quantization is to approximate gradient of the non-differentiable quantizer. To solve this problem, some works have studied relaxed quantization~\cite{louizos2018relaxed, wang2018two, zhuang2018towards, bai2018proxquant}. 
Moreover, with the popularity of automatic machine learning, some recent literature employs reinforcement learning to search for the optimal bitwidth for each layer~\cite{chen2018joint, wu2018mixed, wang2018haq}.
BNNs~\cite{hubara2016binarized, rastegari2016xnor} constrain both weights and activations to binary values (\ie, $+1$  or  $-1$), which brings great benefits to specialized hardware devices. The development of BNNs can be classified into two categories: (i) a focus on improving the training of BNNs~\cite{rastegari2016xnor, tang2017train, hou2017loss, Liu_2018_ECCV}; (ii) multiple binarizations to approximate the full-precision tensor or structure ~\cite{lin2017towards, li2017performance, guo2017network, tang2017train, liu2019circulant, zhuang2019structured}. In this paper, we propose a general auxiliary learning approach that can work on all categories of quantization approaches.

\noindent\textbf{Weight sharing.}	
Weight sharing has been attracting increasing attention for efficient, yet accurate computation. In visual recognition, region proposal networks (RPN) in Faster-RCNN~\cite{ren2015faster} and Mask-RCNN~\cite{he2017mask} share the same backbone with task-specific networks, which greatly saves testing time.
For neural architecture search, ENAS~\cite{pham2018efficient} allows parameters to be shared among all architectures in the search space, which saves orders of magnitude GPU hours. 
In the network compression field, weight/activation quantization intends to partition the weight/activation distribution into clusters and use the centers of clusters as the possible discrete values. This strategy can be interpreted as a special case of weight sharing.
Different from these approaches, we propose to utilize weight sharing 
for jointly optimizing the full-precision auxiliary module and the original low-precision network to improve the accuracy of the latter quantized model.

\noindent\textbf{Auxiliary supervision.}
One straightforward way of adding auxiliary supervision is introducing additional losses into intermediate layers, which serves to combat the vanishing gradient problem while providing regularization. The effectiveness of additional losses has been demonstrated in some literature, like GoogLeNet \cite{szegedy2015going}, DSN \cite{lee2015deeply}, semantic segmentation \cite{zhao2017pyramid, nekrasov2018fast}, etc.
However, these methods are usually sensitive to the positions and scales of the guidance signals. Knowledge distillation (KD) is initially proposed for model compression, where a powerful wide/deep teacher distills knowledge to a narrow/shallow student to improve its performance~\cite{hinton2015distilling, romero2014fitnets}, which can also be treated as adding auxiliary supervisions. 
In terms of the definition of knowledge to be distilled from the teacher, existing models typically use teacher's class probabilities~\cite{hinton2015distilling} and/or intermediate features~\cite{romero2014fitnets, zhuang2018towards, Park_2019_CVPR, Lee_2018_ECCV, zagoruyko2016paying}.
It is worth noting that our proposed auxiliary learning strategy uses weight sharing to assist optimization, where the motivation is very different from the KD methods.
We does not need to pre-train a teacher network which is usually much deeper and may be the upper bound of the performance. Moreover, on network quantization, we show consistent superior performance over KD methods in Sec. \ref{exp:classification}.

\noindent\textbf{Object detection.}
Object detection can be divided into two categories. As one of the dominant detection framework, two-stage detection methods~\cite{girshick2015fast, girshick2014rich, ren2015faster} first generate region proposals and then refine them by subsequent networks.
Another main category is the one-stage methods which are represented by YOLO~\cite{redmon2016you, redmon2017yolo9000, redmon2018yolov3}, SSD~\cite{liu2016ssd} and RetinaNet~\cite{lin2017focal}. The objective is to improve the detection efficiency by directly classifying and regressing the pre-defined anchors without the proposal generation step.
The recent developing trends in object detection is designing light-weight frameworks for mobile applications \cite{chen2017learning, wei2018quantization, tan2019mnasnet}, which usually requires real-time, low-power and fully embedded.
In this paper, we explore to compress and accelerate detectors from the quantization perspective.
Note that, 
we are the first to achieve near lossless 4-bit detectors in the literature.

\section{Method}

In this section, we describe the proposed learning strategy for training a low-precision network. We first give an overview of the proposed approach in Sec.~\ref{method:overview}. Then we introduce the auxiliary module design in Sec.~\ref{method:module} and optimization in Sec.~\ref{method:optimization}, respectively. After that, we provide discussions in Sec. \ref{sec:discussion} and explain how to extend it to quantized object detection in Sec. \ref{method:detection}. Through the following part, we adopt the following terminologies: A layer is a standard parameterized layer in a network such as a dense or convolutional layer. A block is a collection of layers in which the output of its last layer is connected to the input of the next block (\eg, a residual block).

\subsection{Overview and motivation} \label{method:overview}
The overview of the framework is shown in Fig.~\ref{fig:framework}. The blue part, denoted as $F$, shows the low-precision network we aim to learn. The pink part, denoted as $H$, is a full-precision sub-network, which we call the auxiliary module. It connects to intermediate outputs from $F$. The input image is fed into $F$ but generates two outputs, one is from the last layer in $F$ and the other is from the last layer in $H$. In other words, the combination of $F$ and $H$ forms an augmented mix-precision network $F\circ H$ and the parameters of the low-precision network is shared by this augmented network. At the training time, two loss functions are applied to both outputs, and the mixed-precision network and low-precision network are trained jointly. After training, the auxiliary module $H$ will be discarded and only $F$ will be used at the test time.

The motivation of such a design is to create full-precision routes to update parameters of the low-precision model and thus alleviating the difficulty of propagating gradient in a quantized model. Specifically, the intermediate output of each block in $F$ can directly influence the output of the mixed-precision network $F\circ H$ through the full-precision connections in $H$. Consequently, the gradient from the loss of the second output will back propagate to the parameters of each block in the low-precision model.

\subsection{Module design} \label{method:module}

We now elaborate the design for the auxiliary module $H$, which is made up of a sequential of adaptors and aggregators as shown in Fig. \ref{fig:framework} (a).
In particular, the auxiliary module $H$ receives $P$ output feature maps $\{{{\bf{O}}_p}\} _{p = 1}^P$ of the corresponding blocks in $F$.
Let $\{ {B_1},...,{B_P}\}$ be the block indexes where we generate the feature maps.
For the $p$-th input of $H$, we adopt a trainable adaptor $\phi_p ( \cdot )$, which receives the output feature map ${\bf{O}}_p$ of the $B_p$-th block from $F$ and outputs an adapted feature representation
${\phi _p}({{\bf{O}}_p})$.
The motivation of using the adapter is to compensate the distribution discrepancy between the low-precision model and full-precision model. It ensures the quantized activations $\{{{\bf{O}}_p}\} _{p = 1}^P$ to be compatible to the full-precision calculation in $H$. We implement those adapters by a simple $1\times1$ convolutional layer followed by a batch normalization layer in this paper.
In the auxiliary module $H$, the outputs of the adaptor are then sequentially aggregated. Formally, let $\bg_{p}$ denotes the $p$-th aggregated feature. It is achieved by adding the adapted feature ${\phi _p}({{\bf{O}}_p})$ from $F$ and the $(p-1)$-th aggregated feature from $H$ followed by a $\rm{ReLU(\cdot)}$ nonlinearity:
\begin{equation} \label{eq:adapt}
  {\bg_p} = {\rm{ReLU}}(\phi_p ({{\bf{O}}_p}) + {\bg_{p - 1}}). 
\end{equation}
At the last layer in $H$, a classifier layer is applied to ${\bg}_P$ to make the class prediction. Then an auxiliary loss is employed. Note that the auxiliary module $H$ is akin the skip connections in ResNet \cite{he2016deep}.

\subsection{Optimization} \label{method:optimization}

Let $\{ {{\bm{x}}_i},{{\bm{y}}_i}\} _{i = 1}^N$ be the training samples. The proposed method jointly optimize the main network $F$ and the mixed-precision network which is the combination of $F$ and $H$, denoting as $F \circ H$. The training objective is:
\begin{equation} \label{eq:obj}
\begin{aligned}
\mathop {\min }\limits_{\{ {\btheta ^F},{\btheta ^H}\} }\sum\limits_{i = 1}^N & {\cal L}(F({\bx_i};{\btheta ^F}),\by_i ) \\
&+ {{\cal L}_{aux}}((F\circ H)(\bx_i;{\btheta ^H},{\btheta ^F}),\by_i ),
\end{aligned}
\end{equation}
where $\btheta^{F}$ and $\btheta^{H}$ represent the parameters for the backbone $F$ and the auxiliary module $H$, respectively. $\mL$ is the task objective and $\mL_{aux}$ is the auxiliary loss. 
In the classification task, both terms are set to the cross-entropy loss.
From Eq. (\ref{eq:obj}), we can note that $\btheta^{F}$ is shared among $F$ and $F \circ H$.
Following the chain rule, the gradient of $\btheta^{F}$ will have an additional term comes from ${\cal L}_{aux}$. As a result, the approximated gradient is averaged from both the mixed-precision network and the original low-precision network to achieve more accurate updating direction.
In other words, the full-precision module $H$ provides direct gradient for $F$ using weight sharing during back-propagation. %
We summarize the proposed learning process for a quantized neural network in Algorithm~\ref{algo:auxiliary}. 

	\begin{algorithm}[t!]
		\KwIn{Current mini-batch $\{{\bx_i},\by_i\}$; parameter $\btheta^{F}$ of the low-precision network $F$; parameter $\btheta^{H}$ of the full-precision auxiliary module $H$.
		}
		\KwOut{Updated parameters $\{\btheta^{F},\btheta^{H}\}$.}
		
		Obtain the quantized weight $\bQ^{F} = q(\btheta^{F})$, where $q(\cdot)$ is the quantization function; \\
		
		${{\by}_F}, {\by}_H = \mathrm{Forward}({\bx}_i,{{\bQ}^F}, \btheta^{H})$; \\
		
		Compute the loss $\cL({\by_i},{\by}_F)$ for the main network $F$; \\
		
        Compute the loss $\cL_{aux}({\by}_i,{\by}_H)$ for the auxiliary module $H$; \\
        
        $\frac{{\partial {\cL}}}{{\partial {\bQ}^F}}, \frac{{\partial {\cL}_{aux}}}{{\partial \bQ^F}}, \frac{{\partial {\cL}_{aux}}}{{\partial \btheta^H}} = \mathrm{Backward}(\frac{{\partial {\cL}}}{{\partial \by_F}}, \frac{{\partial {\cL_{aux}}}}{{\partial {\by}_H}},{\bQ}^F, \btheta^H$); \\
        
        Compute the gradient, in particular $\nabla {\bQ^F} = \frac{1}{2}(\frac{{\partial \cL}}{{\partial {\bQ^F}}} + \frac{{\partial \cL_{aux}}}{{\partial {\bQ^F}}})$;
        \\
        
		Update parameters using Adam; \\
		
		\caption{\small Joint training approach w.r.t.\
		the main low-precision network $F$ and the full-precision auxiliary module $H$.}
		\label{algo:auxiliary}
	\end{algorithm}

\subsection{Relationship to other methods} 
\label{sec:discussion}
In this section, we will elaborate the relationship between the proposed auxiliary learning and other related approaches.

\begin{figure}[hbt!]
	\centering
	\resizebox{1.0\linewidth}{!}
	{
		\begin{tabular}{c}
			\includegraphics{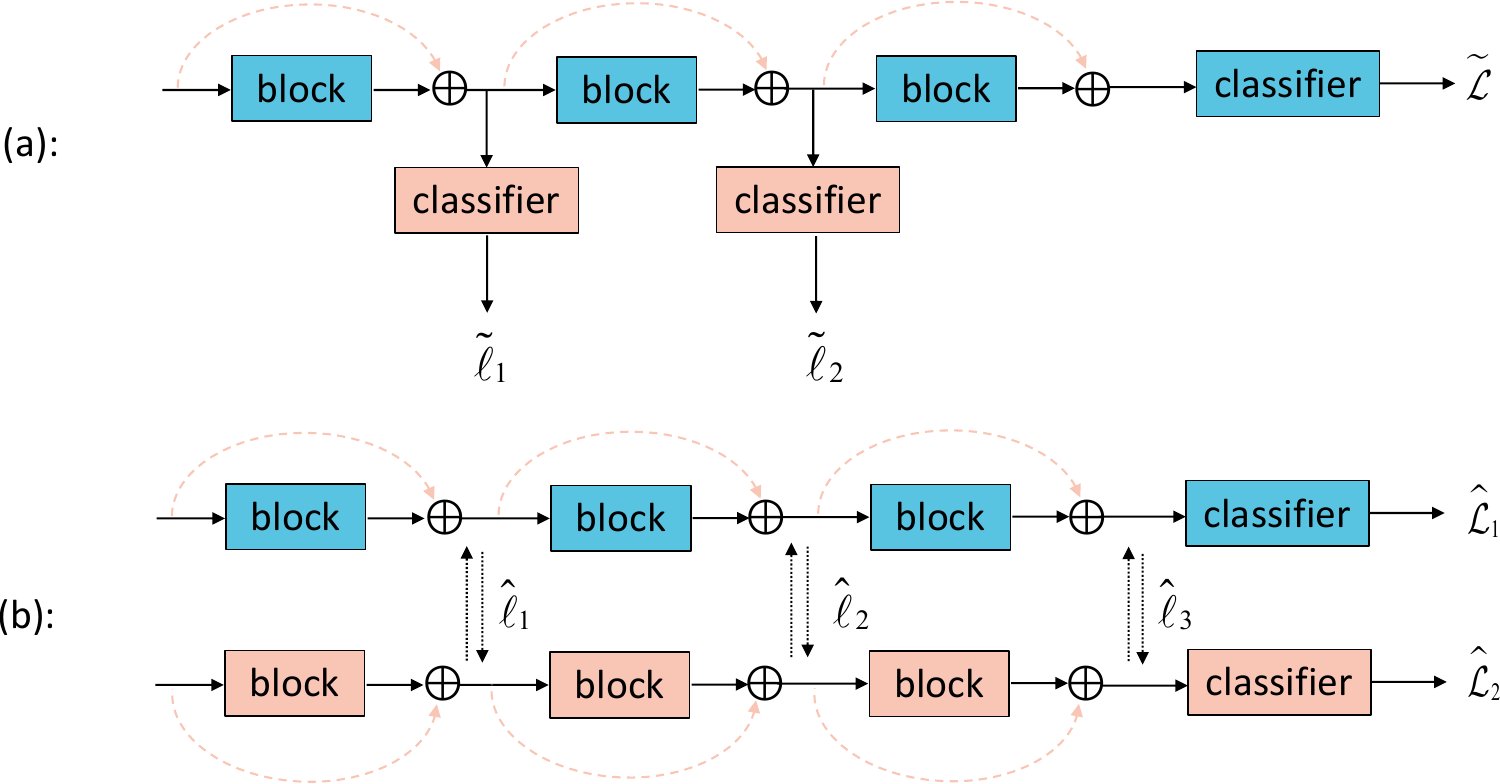}
		\end{tabular}
	}
	\caption{Overview of the two related approaches. (a): Adding additional classification losses to intermediate layers. (b): Knowledge distillation. In network quantization, the pink teacher network is full-precision while the blue student network is low-precision.}
	\label{fig:related}
	\vspace{-1em}
\end{figure}

\noindent\textbf{Differences to applying auxiliary classification losses to intermediate activations.}
An alternatively way to create auxiliary full-precision routes for learning low-precision network is to apply classification losses to intermediate activations. The schematic  illustration of this idea is shown in Fig.~\ref{fig:related} (a), where each intermediate output is attached to a classifier to perform classification.  In such a case, the final training objective becomes:
\begin{equation}
\setlength{\abovedisplayskip}{3pt} 
\setlength{\belowdisplayskip}{3pt}
   {\widetilde \cL_{obj}} = \widetilde\cL + \sum\limits_{i = 1}^M {{\alpha _i}{\widetilde{\ell}_i}}, 
\end{equation}where $\widetilde\cL$ is the classification loss of the original low-precision network, $\widetilde{\ell}_i$ is the classification loss applies to the $i$-th intermediate output, and $\alpha_i$ is the weight associated to the $i$-th loss function. 
This scheme can directly propagate gradient to each block during training through the full-precision classifiers. However, its supervision is very restrictive since it essentially assumes that the intermediate output can be directly used for classification. In practice, we often find that choosing the positions of adding the additional supervisions or the weight $\alpha_i$ can be challenging. An inappropriate setting of those factors may lead to inferior performance than that achieved by directly training the low-precision model.

\noindent\textbf{Differences with knowledge distillation.}
Knowledge distillation (KD) has been explored to assist the quantized model training~\cite{mishra2018apprentice, zhuang2018towards, zhuang2019effective}. In particular, a low-precision student network learns to generate similar posterior probabilities and/or feature representations of a full-precision teacher network (see Fig.~\ref{fig:related} (b)). The training objective can be formulated as  
\begin{equation}
\setlength{\abovedisplayskip}{3pt} 
\setlength{\belowdisplayskip}{3pt}
{\hat{\cL}}_{obj} = {\hat \cL_1} + {\hat \cL_2} + \sum\limits_{i = 1}^M {{\beta _i}{\hat{\ell}_{i}}} ,
\end{equation}
where ${\hat \cL_1}$ and ${\hat \cL_2}$ are task-specific objectives for student and teacher networks respectively. ${\hat{\ell}_{i}}$ represents the $i$-th distillation loss. 

Although both the proposed method and \emph{KD} use a full-precision network to guide the training of a low-precision network, the ways of exerting this guidance are significantly different in \emph{KD} and the proposed method. Specifically, in \emph{KD}, the guidance from the full-precision model to the low-precision model is from the distillation losses while in our method this is achieved by making the parameters of a low-precision model shared with the mixed-precision model. There are many advantages of our method comparing to \emph{KD}: (1) Our method only needs an extra memory to store the auxiliary module rather than a full-precision network. Comparing to \emph{KD}, our method is more memory-efficient. (2) Our method only uses a single auxiliary loss but can create guidance signals to various blocks of the low-precision model. In contrast, \emph{KD} needs multiple distillation losses to achieve this. Thus it usually involves more hyper-parameters, i.e., the weight of each loss term $\beta_i$. Moreover, we empirically find that the proposed learning strategy performs consistently superior than \emph{KD} for learning quantized networks in Sec. \ref{exp:quantize_compare_related} and Sec. \ref{exp:quantize_plain}.

\section{Extension to object detection} 
\label{method:detection} 

Most existing methods evaluate low-precision network with the classification task. Building a low-precision network for more difficult object detection task remains a challenge.
To fill this gap, we further extend our method to build a quantized object detector. 
Following the work in \cite{lin2017focal, lin2017feature}, we consider the object detection framework consisting of a backbone, a feature pyramid and prediction heads. We directly use the quantized network pretrained on the ImageNet classification task to initialize the detection backbone. We adopt the uniform quantization approach QIL \cite{jung2019learning} to quantize both weights and activations, where the quantization intervals are explicitly parameterized and jointly optimized with the network parameters.
We add an individual auxiliary module for each prediction head while sharing a single module for the backbone. 

Beside of applying the proposed auxiliary module and learning strategy to help training the quantized detector, we also propose a modification which we find beneficial. Specifically, unlike \cite{li2019fully} which freezes the batch normalization (BN) statistics during training to stabilize optimization, we instead propose an alternative strategy where the BN statistics still keeps updating: we propose that except the last layers for classification and regression, parameters of the prediction heads are not shared across all feature pyramid levels. This is different from the common full-precision setup. The motivation of this design is that the multi-scale semantic information can not be encoded effectively due to the quantization process at different pyramid levels. For a full-precision network, using a shared head is sufficient to represent rich semantic information for classification and regression with the continuous activations.  However, 
in the low-precision setting, the representational capability of activations is highly degraded due to its discrete values. For the same reason, the batch statistics of quantized activations may differ drastically across different levels.
Therefore, each head should learn independent parameters to capture the corresponding multi-scale information.

\noindent\textbf{Remark:} (1) The proposed modification does not share prediction heads and thus uses more parameters in the low-precision model. However, we should note that without sharing does not increase any additional computational complexity.  Even though the number of parameters is increased, the memory consumption is still significantly reduced comparing to the full-precision model due to the low-bit storage. 
(2) We empirically find that not sharing heads may not improve (but reduce) the performance in the full-precision setting. So the proposed modification is only for low-precision networks. Please check the experiments in Sec. \ref{exp:detection_ablation} for more discussions. 

\section{Experiments} \label{sec:exp}
In this section, we evaluate our proposed methods on image classification in Sec. \ref{exp:classification} and object detection in Sec. \ref{exp:detection}, respectively.
To investigate the effectiveness of the proposed method, we define several methods for comparison: 
\textbf{Auxi:} We optimize the network with the auxiliary module. 
\textbf{KD:} We employ the joint knowledge distillation in \cite{zhuang2019effective, zhuang2018towards} to improve the quantized network.
\textbf{Additional loss:} We evenly insert classification losses at intermediate layers to assist training. Note that we will detail the settings in specific sections.

\subsection{Experiments on image classification}  \label{exp:classification}

We perform experiments on two standard image classification datasets: CIFAR-100 \cite{krizhevsky2009learning} and ImageNet \cite{russakovsky2015imagenet}. The CIFAR-100 dataset consists of 60,000 color images of size $32 \times 32$ belonging to 100 classes. There are 50,000 training and 10,000 test images. ImageNet contains about 1.2 million training and 50K validation images of 1,000 object categories. 
To verify the effectiveness of the proposed auxiliary learning strategy, we experiment on various representative quantization approaches, including uniform fixed-point approach DoReFa-Net~\cite{zhou2016dorefa}, non-uniform fixed-point method LQ-Net~\cite{zhang2018lq}, as well as binary neural network approaches BiReal-Net~\cite{Liu_2018_ECCV} and Group-Net~\cite{zhuang2019structured}. 

\vspace{-2mm}
\subsubsection{Implementation details}
Following previous approaches~\cite{zhang2018lq, zhou2016dorefa, hubara2016binarized, zhuang2018towards, zhu2016trained}, we quantize all the convolutional layers to ultra-low precision except the first and last layers. However, to further improve the efficiency, we quantize the first convolutional layer and the last fully-connected layer to 8-bit.
We first pre-train the full-precision counterpart as initialization and then fine-tune the quantized model. For all ImageNet experiments, training images are resized to $256 \times 256$, and $224 \times 224$ patches are randomly cropped from an image or its horizontal flip, with the per-pixel mean subtracted. We use the single-crop setting for testing. No bias terms are used. We use SGD optimizer for the pre-training stage. For the fine-tuning stage, we adopt the Adam optimizer~\cite{kingma2014adam}. The mini-batch size is set to 256. We train a maximum 35 epochs and decay the learning rate by 10 at the 25-th and 30-th epochs. For fine-tuning the fixed-point methods~\cite{zhou2016dorefa, zhang2018lq}, the learning rate is initialized to 1e-3. For fine-tuning binary neural networks~\cite{Liu_2018_ECCV, zhuang2019structured}, the initial learning rate is set to 5e-4. In practice, we take the output of each residual block ~\cite{he2016deep} as the input of the auxiliary module. 
Our implementation is based on PyTorch.

\subsubsection{Effect of the auxiliary module}  \label{exp:quantize_auxiliary}

\vspace{-2mm}
\begin{table}[hbt!]
	\centering
	\caption{Accuracy (\%) of different comparing methods on the ImageNet validation set.}
	\scalebox{0.85}
	{
		\begin{tabu}{c | c | c c}
		    model &	method &Top-1 acc. &Top-5 acc. \\
			\tabucline[1pt]{-}
			\multirow{2}{*}{ResNet-101} & DoReFa-Net (2-bit)  &70.8  &89.6  \\ 
			& DoReFa-Net + Auxi  &\bf{74.6}   &\bf{91.9}    \\
			\hline			
			\multirow{2}{*}{ResNet-50} & DoReFa-Net (2-bit)  &70.2  &89.1  \\ 
			& DoReFa-Net + Auxi  &\bf{73.8}  &\bf{91.4} \\
			\hline
		   \multirow{2}{*}{ResNet-50} & LQ-Net (3-bit)  &74.2  &91.6 \\
			& LQ-Net + Auxi &\bf{75.4}  &\bf{92.4} \\
			\hline 
			\multirow{2}{*}{ResNet-18} & BiReal-Net &56.4  &79.5 \\
			& BiReal-Net + Auxi & \bf{58.6}  &\bf{81.2}  \\
			\hline
			\multirow{2}{*}{ResNet-18} & Group-Net (5 bases) &64.8  &85.7 \\
			& Group-Net + Auxi &\bf{66.0}  &\bf{86.5} \\
			
	\end{tabu}}
    \vspace{-1em}
	\label{tab:with_shortcut}
\end{table}

\begin{figure}[bt!]
	\centering
	\resizebox{0.785\linewidth}{!}
	{
		\begin{tabular}{c}
			\includegraphics{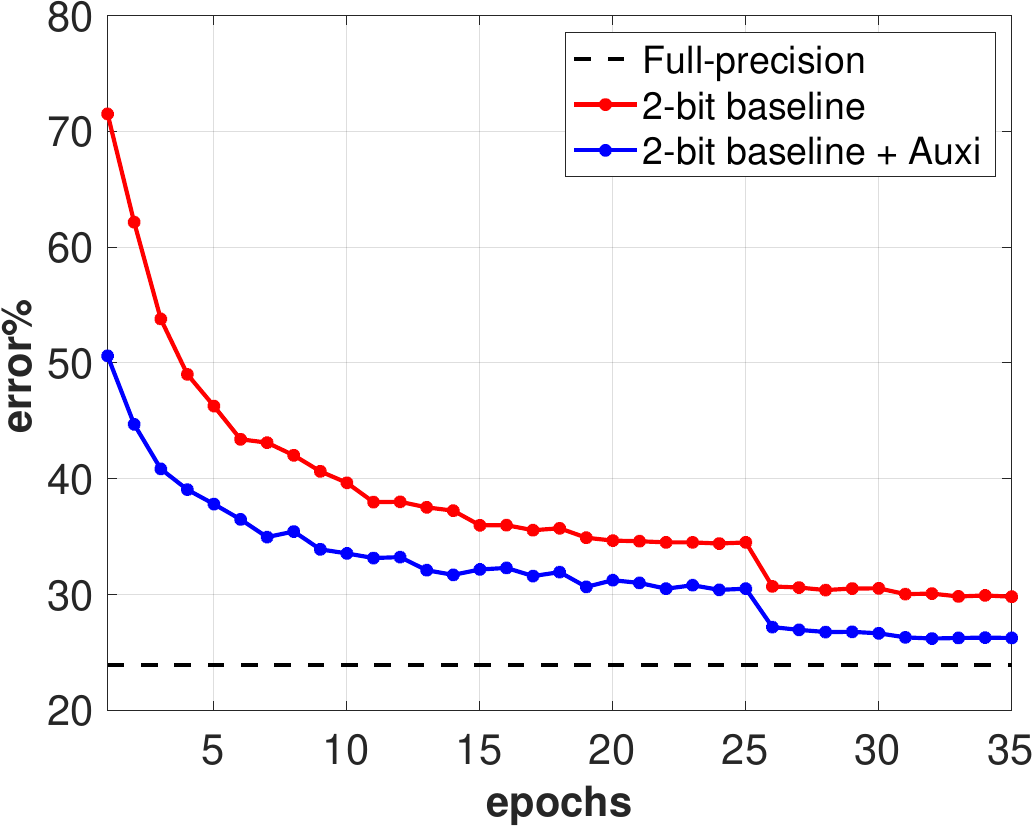}
		\end{tabular}
	}
	\caption{The convergence curves of 2-bit DoReFa-Net ResNet-50 baseline and the proposed auxiliary learning approach on ImageNet validation set during fine-tuning.}
	\label{fig:quantization}
	    \vspace{-1em}
\end{figure}

In this section, we explore the effect of auxiliary module on assisting the low-precision network optimization. 
The results are reported in Table~\ref{tab:with_shortcut}. By combining the baseline with \emph{Auxi}, we can observe a steady performance increase compared with the original baseline. This strongly supports that the learned auxiliary gradient can facilitate the convergence of the low-precision model. In particular, the gradient of the shared weights is averaged from both the auxiliary module and the original low-precision network to achieve more accurate updating direction. Moreover, increasing gradient paths is important to solve the non-differentiability of the discrete quantization process has been proved by~\cite{bethge2018training, Liu_2018_ECCV}. %
To make it more clear, we plot the convergence curves of DoReFa-Net with ResNet-50 in Fig.~\ref{fig:quantization}. From the figure, we can observe that \emph{baseline + Auxi} converges much faster and better than the baseline. After the first epoch, the \emph{baseline + Auxi} outperforms \emph{baseline} by $\sim20$\% on the Top-1 accuracy. This result strongly justifies that the auxiliary module efficiently solves the non-differentiable problem by providing accurate hierarchical gradient for updating parameters during back-propagation.

It is worth noting that when the network becomes deeper (e.g., ResNet-50, 101), the improvement turns out to be more obvious. For instance, \emph{Auxi} brings \textbf{3.8}\% Top-1 accuracy increase over the 2-bit baseline on ResNet-101. It can be attributed to that when the quantized network goes deeper, the optimization becomes more difficult due to the non-differentiable discretization process. However, the full-precision auxiliary module can provide direct hierarchical gradient to effectively solve this problem.
Note that with the basic uniform DoReFa-Net baseline, we currently achieve the comparing results with state-of-the-arts on ResNet-50 and ResNet-101 \cite{zhang2018lq, louizos2018relaxed, jung2019learning} without advanced non-uniform or relaxation strategies. 

\vspace{-1em}
\subsubsection{Comparison with other related approaches} \label{exp:quantize_compare_related}

In this section, we compare the auxiliary module with the related approaches discussed in Sec.~\ref{sec:discussion} and report the performance in Table~\ref{tab:exp_compare_related}.
The experiments are based on the 2-bit DoReFa-Net with ResNet-18, ResNet-34 and ResNet-50 on ImageNet. For KD experiments, the results are directly cited from \cite{zhuang2019effective}.
We can observe that introducing additional losses into intermediate layers does not show obvious improvement to the final performance.
Compared to \emph{KD}, we don't need to pre-train a complex teacher network whose quality is sensitive to the final performance. In contrast, we propose a simpler yet effective weight sharing strategy to jointly optimize the low-precision network and the floating-point auxiliary module. More detailed analysis on the difference between two approaches can be referred to Sec. \ref{sec:discussion}.
From the results, we can observe that \emph{Auxi} consistently outperforms \emph{KD}. For instance, on ResNet-50, \emph{Auxi}  
exceeds \emph{KD} by $\textbf{2.4}$\% on the Top-1 accuracy.
These results justify that the auxiliary module can effectively solve the non-differentiable problem during back-propagation in the low-precision network training. With a very different learning strategy, \emph{Auxi} shows consistently superior empirical results than \emph{KD}. We therefore argue that the proposed auxiliary learning may be a substitute to \emph{KD} methods  \cite{zhuang2018towards, mishra2018apprentice} in network quantization. 

\begin{table}[hbt!] 
	\centering
	\caption{Accuracy (\%) of different supervision strategies on the ImageNet validation set based on 2-bit DoReFa-Net on ResNet-18, ResNet-34 and ResNet-50.}
	\scalebox{0.85}
	{
		\begin{tabu}{c | c |c c}
		    model &	method &Top-1 acc. &Top-5 acc.\\
			\tabucline[1pt]{-}
			\multirow{4}{*}{ResNet-18} & baseline (2-bit) &64.7  &86.0   \\ 
			&baseline + Additional loss  &64.9   &86.1   \\
			&baseline + KD &65.6   &86.3  \\ 
			&baseline + Auxi &\bf{66.7}   &\bf{87.0}   \\
			\hline
			\multirow{4}{*}{ResNet-34} & baseline (2-bit) &68.2   &88.1  \\ 
			&baseline + Additional loss   &68.5  &88.2 \\
			&baseline + KD  &69.0  &88.6   \\ 
			&baseline + Auxi &\bf{71.2}   &\bf{89.8}   \\
            \hline
			\multirow{4}{*}{ResNet-50} & baseline (2-bit) &70.2  &89.1  \\ 
			&baseline + Additional loss  &70.5   &89.3  \\
			&baseline + KD &71.4  &90.0   \\ 
			&baseline + Auxi &\bf{73.8}  &\bf{91.4}   \\

	\end{tabu}}
	\vspace{-1em}
	\label{tab:exp_compare_related}
\end{table}

\subsubsection{Experiments on plain networks}   
\label{exp:quantize_plain}

\begin{table}[hbt!] 
	\centering
	\caption{Accuracy (\%) of the proposed approaches on the ImageNet validation set. All the cases are 2-bit and without skip connections except for the baselines.
	We can observe that the auxiliary module can significantly improve the plain network performance.}
	\scalebox{0.78}
	{
		\begin{tabu}{c | c | c c}
		    model &	method &Top-1 acc. &Top-5 acc. \\
			\tabucline[1pt]{-}
			\multirow{4}{*}{DoReFa-Net on ResNet-18} & baseline (2-bit) &64.7  &86.0  \\
			&plain &61.5  &84.3 \\
			&plain + KD &62.7  &85.0 \\
			&plain + Auxi &\bf{63.9}  &\bf{85.5} \\
			\hline
			\multirow{4}{*}{DoReFa-Net on ResNet-34} & baseline (2-bit) &68.2  &88.1  \\
			 &plain &62.1  &83.9  \\
			 &plain + KD &64.5  &85.4  \\ 
			&plain + Auxi &\bf{66.4}  &\bf{86.8}  \\
			\hline
		  \multirow{4}{*}{LQ-Net on ResNet-34} &baseline (2-bit) &69.8 &89.1 \\
			&plain  &63.5  &84.6 \\
			&plain + KD &65.7   &86.8  \\
			&plain + Auxi &\bf{68.6}  &\bf{88.5} \\
			
	\end{tabu}}
	\vspace{-1.0em}
	\label{tab:plain_imagenet}
\end{table}

 \begin{table}[hbt!] 
	\centering
	\caption{Accuracy (\%) of 2-bit DoReFa-Net using ResNet-18 on the CIFAR-100 dataset.}
	\scalebox{0.9}
	{
		\begin{tabu}{c | c | c c}
		    model &	method &Top-1 acc. &Top-5 acc. \\
			\tabucline[1pt]{-}
			\multirow{4}{*}{ResNet-18} 
			& full-precision &70.7  &91.3 \\
			\cline{2-4}
			& baseline (2-bit) &67.6  &\bf{90.2} \\
			& plain &64.6  &88.3   \\ 
			& plain + Auxi  &\bf{67.9}  &90.0 \\
	\end{tabu}}
	\vspace{-1mm}
	\label{tab:plain_cifar100}
\end{table}

We further explore an interesting by-product of the auxiliary module for network quantization. We assume that the auxiliary module mimics the effect of skip connections and can partially share its effect.
We therefore analyze training a plain low-precision network without skip connections. The results can be referred in Table \ref{tab:plain_imagenet} and Table~\ref{tab:plain_cifar100}. 
\emph{plain} represents we directly optimize a low-precision plain network without skip connections. 
By comparing \emph{plain} and \emph{plain + Auxi}, we observe apparent accuracy increase by incorporating \emph{Auxi}.
For example, in LQ-Net ResNet-34 based experiments, introducing \emph{Auxi} can boost the Top-1 accuracy by $5.1$\%. On tiny CIFAR-100 dataset, \emph{plain + Auxi} even outperforms the Top-1 baseline. Moreover, same as the observation in Sec. \ref{exp:quantize_compare_related},  \emph{Auxi} still performs consistently better than \emph{KD}. From the plain network setting, we can strongly justify that the auxiliary module can provide hierarchical gradient to promote convergence of the quantized network.

However, we still observe performance gap between \emph{plain + Auxi} and the baseline on large-scale ImageNet. This can be attributed to two assumptions of skip connections. First, the skip connections may improve the convergence of training, as indicated by the improvement observed when using \emph{Auxi}.
Second, the skip connection and the feature map after one convolution are added through a tensor addition. Then the representational capability (i.e., the value range) of each entry in the added activations is significantly enhanced. In other words, the plain network has less representational capability than its residual counterpart.

\vspace{-0.5em}
\subsubsection{Effect of different auxiliary architectures}    \label{exp:quantize_archi}

We further explore the influence of different auxiliary module architectures in Table \ref{tab:quantize_archi}.
From the table, we observe that increasing the complexity of the auxiliary module can further boost the performance. For example, by replacing the $1\times1$ convolution in the adaptor with a larger kernel of $3\times3$, we further get slightly performance gain. This can be attributed to that the gradient of shared parameters is averaged from $F$ and $F \circ H$, where better representational capability of $H$ can result in more accurate gradient update.

\begin{table}[hbt!] 
	\centering
	\caption{Accuracy (\%) of using different adaptors. We use DoReFa-Net on ImageNet as our baseline.}
	\scalebox{0.9}
	{
		\begin{tabu}{c | c | c c}
		    model &	method &Top-1 acc. &Top-5 acc. \\
			\tabucline[1pt]{-}
			\multirow{3}{*}{ResNet-18} & baseline (2-bit) &64.7   &86.0   \\
			&baseline + $1 \times 1$ Auxi & 66.7  & 87.0 \\
			&baseline + $3 \times 3$ Auxi & \bf{66.9}  & \bf{87.1} \\
	
	\end{tabu}}
	\label{tab:quantize_archi}
	\vspace{-2mm}
\end{table}

\subsection{Experiments on quantized object detection} \label{exp:detection}

In this section, we evaluate the proposed approach on the general object detection task. Our experiments are conducted on the large-scale detection benchmark COCO~\cite{lin2014microsoft}. Following~\cite{lin2017feature, lin2017focal}, we use the COCO \textit{trainval35k} split (115K images) for training and \textit{minival} split (5K images) for validation. We conduct experiments based on RetinaNet \cite{lin2017focal} and compare with the state-of-the-art FQN~\cite{li2019fully}.

\subsubsection{Training details}
Training is divided into two stages. In the first stage, the detection framework is kept to full-precision. The backbone is initialized with the classification model pre-trained on the ImageNet dataset. Unless specified, we use the same hyper-parameters with RetinaNet. Specifically, all training and evaluation images are resized so that their shorter edges are 800 pixels. We augment training images by random horizontal flipping while no evaluation augmentations are performed. 
Our network is trained with stochastic gradient descent (SGD) for 90K iterations with the initial learning rate being 0.01 and the batch size of 16. The learning rate is decayed by a factor of 10 at iterations 60K and 80K, respectively. 
In the second stage, we use the converged model in the first stage as initialization and fine-tune with quantization. This stage uses identical settings as the full-precision training, except that we use Adam optimizer and the initial learning rate is set to 1e-3. Our implementation is based on Detectron2 \cite{wu2019detectron2}.

\vspace{-2mm}
\subsubsection{Performance evaluation}

We report the performance of the proposed quantized detection framework in Table \ref{tab:detection_val}. From the results, we can observe that our 4-bit detector can achieve near lossless results over the full-precision counterparts, which meets the requirement for practical
deployment. Moreover, we can achieve significant performance boost over FQN on all comparing architectures. For example, on ResNet-50, the improvement reaches to $3.6$ on AP.

\begin{table}[hbt!]
	\centering
	\caption{Performance on the COCO validation set with 4-bit quantization.}
	\scalebox{0.8}{
	\begin{tabular}{c | c |c c c | c c c}
	     Backbone & Method  &AP &AP$_{50}$  &AP$_{75}$   &AP$_{S}$  &AP$_{M}$  &AP$_{L}$ \\\hline
		\multirow{3}{*}{ResNet-50} & RetinaNet  &36.5  &56.5  &39.2   &21.4   &40.4   &46.9   \\
		& FQN~\cite{li2019fully}  &32.5  &51.5 &34.7  &17.3 &35.6 &42.6 \\
		& Ours   &\bf{36.1}  &\bf{55.8}   &\bf{38.9}   &\bf{21.2}    &\bf{39.9}  &\bf{46.3}    \\\hline
		\multirow{3}{*}{ResNet-34} & RetinaNet &35.2  &54.3  &37.6  &19.5  &38.5  &46.2  \\
		& FQN~\cite{li2019fully} &31.3 &50.4 &33.3 &16.1 &34.4 &41.6 \\
		& Ours  &\bf{34.7}  &\bf{53.7}  &\bf{36.9}  &\bf{19.3}  &\bf{38.0}  &\bf{45.9} \\ \hline
		\multirow{3}{*}{ResNet-18} &RetinaNet &32.1  &50.5  &34.1  &16.9  &34.8  &42.6  \\ 
		&FQN~\cite{li2019fully}  &28.6 &46.9 &29.9  &14.9 &31.2 &38.7   \\
		&Ours  &\bf{31.9}  &\bf{50.4} &\bf{33.7}  &\bf{16.5}  &\bf{34.6}  &\bf{42.3} \\
		
	\end{tabular}}
	\label{tab:detection_val}
	\vspace{-2em}
\end{table}

\subsubsection{Ablation studies}
\label{exp:detection_ablation}

We now perform ablation studies to give comprehensive analysis and insights of the quantized detection framework. The results are reported in Table \ref{tab:detection_ablation}. ``Backbone only'' indicates we only quantize the backbone to 4-bit while other parts are kept to full-precision. ``Baseline'' represents we directly quantize the RetinaNet using QIL \cite{jung2019learning}.  

When we only quantize the backbone network, we don't observe AP drop at least on ResNet-18. This justifies that 4-bit backbone can encode accurate features for further decoding. However, when quantizing all the components including feature pyramid and prediction heads, we can find obvious precision drop. The baseline result indicates that quantizing the continuous features to a fixed range of integers can cause great multi-scale information loss. 

We now explore the effect of the two strategies described in Sec. \ref{method:detection}.
First, we incorporate the proposed auxiliary learning strategy to assist the convergence of quantized detector. Specifically, \emph{Baseline + Auxi} can boost the AP of \emph{Baseline} by $1.4$.
Second, not sharing heads is particularly designed for the low-precision detector. We should note that this strategy deteriorates the performance in the full-precision setting. The reason is that a separate head can only see a certain range size of objects.
In contrast, not sharing heads can boost the AP by $1.3$ in the quantization setting compared with the sharing heads counterpart. 
It can be attributed that when quantizing both weights and activations to 4-bit, the representational capability of each head is very limited and the statistics of activations differ a lot.
Therefore, each head should learn independent parameters to transform the corresponding level feature for classification and regression. 

\begin{table}[bt!]
	\centering
	\caption{Ablation studies on the COCO validation set with 4-bit quantization.}
	\scalebox{0.73}{
	\begin{tabular}{c |c c c | c c c}
		Method &AP &AP$_{50}$  &AP$_{75}$   &AP$_{S}$  &AP$_{M}$  &AP$_{L}$ \\\hline
		RetinaNet (w/ sharing) &32.1  &50.5  &34.1  &16.9  &34.8  &42.6  \\
		RetinaNet (w/o sharing) &31.2  &49.2  &32.9  &15.8  &34.3  &41.5 \\
		\cdashline{1-7}
		Backbone only  (w/ sharing) &32.1  &50.7  &34.0  &16.7  &34.7  &42.7 \\
		Baseline (w/ sharing)  &29.2   &47.1  &31.0  &14.4   &31.5  &38.5  \\
		Baseline + Auxi (w/ sharing)  &30.6  &48.8  &32.8   &15.5   &33.1   &40.2  \\
		Baseline + Auxi (w/o sharing)  &\bf{31.9}  &\bf{50.4} &\bf{33.7}  &\bf{16.5}  &\bf{34.6}  &\bf{42.3}  \\
		
	\end{tabular}}
	\label{tab:detection_ablation}
	\vspace{-2em}
\end{table}
\section{Conclusion}
In this paper, we have proposed an auxiliary learning strategy to tackle the 
non-differentiable quantization process in training low-bitwise convolutional neural networks. 
Specifically, we have explicitly utilized weight sharing to construct a full-precision auxiliary module. During training, the auxiliary module is combined with the low-precision network to form a mix-precision network, which is jointly optimized with the low-precision model.
In this way, the full-precision auxiliary module can provide direct hierarchical gradient during back-propagation to assist the optimization of the low-precision network. In the testing phase, the auxiliary module is removed without introducing any additional computational complexity.
Moreover, we have also worked on quantized object detection and proposed several practical solutions.
We have conducted extensive experiments based on various quantization approaches and observed consistent performance increase on the image classification and object detection. To be emphasized, we have achieved near lossless results using 4-bit detectors.

\textbf{Acknowledgement
}
M. Tan was in part supported by Guangdong Provincial Scientific and Technological Funds under Grants 2018B010107001.
This work was in part supported by ARC DP Project `Deep learning that scales'.

{\small
\bibliographystyle{ieee_fullname}
\bibliography{egbib}
}

\end{document}